\documentclass[journal,onecolumn, draftclsnofoot]{IEEEtran}

\ifCLASSINFOpdf

\else

\fi

\usepackage{cite}
\usepackage{amsmath,amssymb,amsfonts}
\usepackage{algorithmic}
\usepackage{graphicx}
\usepackage{textcomp}

\usepackage{amsmath,amssymb,bm}
\usepackage{amsfonts,color,bbm}
\usepackage{subfig}
\usepackage{algorithm,algorithmic}
\usepackage{amsthm}
\usepackage{adjustbox}

\usepackage{xcolor}
\usepackage{rotating}
\usepackage{multirow}

\DeclareMathOperator{\sgn}{sgn}

\renewcommand{\vec}[1]{\mbox{\boldmath${#1}$}}

\def\BibTeX{{\rm B\kern-.05em{\sc i\kern-.025em b}\kern-.08em
    T\kern-.1667em\lower.7ex\hbox{E}\kern-.125emX}}
    
\begin{document}

\title{A Neural Network Approach for Online Nonlinear Neyman-Pearson Classification}

\author{Basarbatu Can,
        Huseyin Ozkan
\thanks{This work was supported by The Scientific and Technological Research Council of Turkey (TUBITAK) under Contract 118E268.}}


\maketitle

\begin{abstract}
We propose a novel Neyman-Pearson (NP) classifier that is both online and nonlinear as the first time in the literature. The proposed classifier operates on a binary labeled data stream in an online manner, and maximizes the detection power about a user-specified and controllable false positive rate. Our NP classifier is a single hidden layer feedforward neural network (SLFN), which  is initialized with random Fourier features (RFFs) to construct the kernel space of the radial basis function at its hidden layer with sinusoidal activation. Not only does this use of RFFs provide an excellent initialization with great nonlinear modeling capability, but it also exponentially reduces the parameter complexity and compactifies the network to mitigate overfitting while improving the processing efficiency substantially. We sequentially learn the SLFN with stochastic gradient descent updates based on a Lagrangian NP objective. As a result, we obtain an expedited online adaptation and powerful nonlinear Neyman-Pearson modeling. Our algorithm is appropriate for large scale data applications and provides a decent false positive rate controllability with real time processing since it only has $O(N)$ computational and $O(1)$ space complexity ($N$: number of data instances). In our extensive set of experiments on several real datasets, our algorithm is highly superior over the competing state-of-the-art techniques, either by outperforming in terms of the NP classification objective with a comparable computational as well as space complexity or by achieving a comparable performance with significantly lower complexity.
\end{abstract}


\IEEEpeerreviewmaketitle

\section{Introduction}\label{sec:Intro}
\IEEEPARstart{D}{esigning} a binary classifier with asymmetrical costs for the errors of type I (false positive) and type II (false negative) \cite{access1, access2,pamicost}, or equivalently designing a Neyman-Pearson classifier \cite{davenport2010tuning}, is required in various applications ranging from facial age estimation \cite{ageestimation}, multi-view learning \cite{multiviewlearn} and software defect prediction \cite{defect_pred} to video surveillance \cite{saligrama2012video} and data imputation \cite{ozkan2015data}. For example, in medical diagnostics, type II error (misdiagnosing as healthy) has perhaps more severe consequences, whereas type I error (misdiagnosing as unhealthy) may result in devastating psychological effects \cite{jiang2014cost}. In this example, the error costs must be determined probably asymmetrically for cost sensitive learning \cite{access1, access2} of the desired classifier. However, it is often more convenient -but technically equivalent \cite{davenport2010tuning}- to describe the user needs by the maximum tolerable type I error, cf. \cite{tong2016survey} and the references therein, instead of having to determine the error costs to meet the tolerance. This leads to the Neyman-Pearson (NP) characterization of the desired classifier \cite{davenport2010tuning} and false positive rate controllability, where the goal is to maximize the detection power, i.e., minimize type II error, while upper-bounding the false positive rate, i.e., type I error, by a user-specified threshold.

To this goal, as the first time in the literature, we introduce a novel online and nonlinear NP classifier based on a single hidden layer feedforward neural network (SLFN), which is sequentially learned with a Lagrangian non-convex NP objective (i.e. maximum detection power about a controllable user specified false positive rate). We use stochastic gradient descent (SGD) optimization for scalability to voluminous data and online processing with limited memory requirements. During the SGD iterations, we a) sequentially infer the value of the Lagrangian multiplier in a data driven manner to obtain the correspondence between the asymmetrical error costs and the desired type I error rate, and b) update all the SLFN parameters to maximize the detection power (minimize the resulting cost sensitive classification error) at the desired false positive rate. To achieve powerful nonlinear modeling and improve scalability, we use the SLFN in a kernel inspired manner, cf. \cite{Rahimi} for the kernel approach to nonlinearity. For this purpose, the hidden layer is initialized with a sinusoidal activation to approximately construct the high dimensional kernel space (of any symmetric and shift invariant kernel under Mercer's conditions, e.g., radial basis function) through the random Fourier features (RFFs) \cite{Rahimi}. The output layer follows with identity activation.

We emphasize that the kernel inspired SLFN has two benefits: expedited powerful nonlinear modeling and scalability. Namely, first, it enables an excellent network initialization as RFFs are already sufficiently powerful to learn complex nonlinear decision boundaries even when kept untrained. This speeds up and enhances the learning of complex nonlinearities by relieving the burden of network initialization. Second, the hidden layer is compactified thanks to the exponential rate of improvement in approximating the high dimensional kernel space due to Hoeffding's inequality \cite{Rahimi}. As a result, the number of hidden nodes, parameter complexity and the computational complexity of forward-backward network evaluations reduce, and therefore the scalability substantially improves while also mitigating overfitting. Moreover, thanks to the learning of the hidden layer, the randomly initialized Fourier features are continuously improved during SGD steps for even further compactification and better nonlinear modeling. Hence, our online NP classifier is powerfully nonlinear and computationally highly efficient with $O(N)$ processing and negligible $O(1)$ space complexity, where $N$ is the number of data instances.

The main contribution of our work is that we are the first to propose a Neyman-Pearson (NP) classifier that is both online and nonlinear. Our algorithm -as an important novel addition to the literature- is appropriate for contemporary fast streaming large scale data applications that require real time processing with capabilities of complex nonlinear modeling and false positive rate controllability. In our extensive experiments, the introduced classifier yields significantly better results compared to the competing state-of-the-art NP techniques; either performance-wise (in terms of the detection power and false positive rate controllability) at a comparable computational and space complexity, or efficiency-wise (in terms of complexity) at a comparable performance. The presented study is also the first to design a neural network (as an SLFN) in the context of NP characterization of classifiers, which is expected to open up new directions into deeper architectures since the NP approach has been left surprisingly unexplored in deep learning.

In the following Section \ref{sec:RW}, we discuss state-of-the-art NP classification methods. We provide the problem description in Section \ref{sec:Problem Desc}, and then introduce our technique for online and nonlinear NP classification in Section \ref{sec:Method}. After the experimental evaluation is presented in Section \ref{sec:Experiments}, we conclude in Section \ref{sec:Conclusion}.

\section{Related Work}\label{sec:RW}
Neyman-Pearson classification has found a wide-spread use across various applications due to the direct control over the false positive rate that it offers, cf. \cite{tong2016survey} and the references therein. For example, an NP classifier is commonly employed for anomaly detection, where the false positive rate controllability is particularly important. In the one class formulation (due to the extreme rarity of anomalies) of anomaly detection \cite{scott2006learning,hero2007geometric,zhao2009anomaly,saligrama2012local}, the NP classification turns out (when the anomalies are assumed uniformly distributed) estimating the minimum volume set (MVS) that covers $1-\tau$ fraction of the nominal data ($\tau$ is the desired false positive rate). Then, an instance is anomalous if it is not in the MVS. A structural risk minimization approach is presented in \cite{scott2006learning} for learning the MVS based on a class of sets generated by a dyadic tree partitioning. Geometric entropy minimization \cite{hero2007geometric} and empirical scoring \cite{zhao2009anomaly} can also be used to estimate the MVS, both of which are based on nearest neighbor graphs. The scoring of \cite{zhao2009anomaly} is later extended to the local anomaly detection in \cite{saligrama2012local} and a new one class support vector machines (SVM) in \cite{chen2013new}. Although the algorithms in these examples with batch processing, i.e., not online, have decent theoretical performance guarantees, they are not scalable to large scale data due to their prohibitive computational as well as space complexity and hence they cannot be used in our scenario of fast streaming applications. Online extensions to the original batch one class SVM \cite{scholkopf2001estimating}, which can be shown to provide an estimator of the MVS \cite{vert2006consistency}, have been proposed for distributed processing \cite{miao2018distributed} and wireless sensor networks \cite{zhang2009adaptive}. However, neither these online extensions nor the original one class SVM address the false positive rate controllability as they require additional manual parameter tuning for that. In contrast, our proposed online NP classifier directly controls (without parameter tuning) the false positive rate and maximizes the detection power with nonlinear modeling capabilities. Furthermore, NP formulation in the one class setting requires the knowledge of the target density (e.g., anomaly), which is often unknown and thus typically assumed to be uniform; but then the problem can be turned into a supervised binary NP classification by simply sampling from the assumed target density. On the other hand, when there is also data from the target class, the one class formulation in aforementioned studies does not directly address how to incorporate the target data. Hence, our two class supervised formulation of binary NP classification also covers the solution of the one class classification, and our proposed algorithm is consequently more general and applicable in both cases of target data availability.

Among the two class binary NP classification studies (cf. \cite{tong2016survey} for a survey), plug-in approaches (such as \cite{other1} and \cite{tong2013plug}) based on density estimation as an application of the NP lemma \cite{scott2005neyman} are difficult to be applied in high dimension due to overfitting \cite{hero2007geometric}. Particularly, \cite{other1} exploits the expectation-maximization algorithm for density estimation using a neural network with -however- batch processing and manual tuning for finding the threshold to satisfy the NP type I error constraint. In \cite{other5}, a neural network is trained with symmetric error costs for modeling the likelihood ratio, which is thresholded to match the desired false positive rate but determining the threshold requires additional work. Moreover, the approach of thresholding after training with symmetric error costs (cf. \cite{tong2016survey} for other examples in addition to \cite{other5}) does not yield NP optimality, since NP classification requires training with asymmetric error costs corresponding to the desired false positive rate. Unlike our presented work, approaches in \cite{other1,tong2013plug,other5} are also not online and do not allow real time false positive rate controllability. Recall that NP classification is equivalent to cost sensitive learning \cite{davenport2010tuning} when the desired false positive rate can be accurately translated to error costs, but achieving an accurate translation, i.e., correspondence, is typically nontrivial requiring special attention \cite{davenport2010tuning, kong2019false}. This correspondence problem is addressed  i) in \cite{davenport2010tuning} as parameter tuning with improved error estimations, and ii) in \cite{kong2019false} as an optimization with the assumption of class priors and unlabeled data. Besides the exploitation of SVM \cite{davenport2010tuning}, other classifiers such as logistic regression \cite{cox1958regression} have also been considered in \cite{tong2018neyman} and incorporated into a unifying NP framework as an umbrella algorithm. We emphasize that these approaches, the SVM based tuning approach \cite{davenport2010tuning} and the risk minimization of \cite{scott2005neyman} as well as the umbrella algorithm \cite{tong2018neyman} in addition to the optimization of \cite{kong2019false}, do not satisfy our computational online processing requirements, as they are batch techniques and not scalable to large scale data.

In most of the contemporary fast streaming data applications, such as computer vision based surveillance \cite{lu2013abnormal} and time series analysis \cite{ozkan2015online}, computationally efficient processing along with only limited space needs is a crucial design requirement. This is necessary for scalability in such applications which constantly generate voluminous data at unprecedented rates. However, the literature about the Neyman-Pearson classification (cf. \cite{tong2016survey} for the current state) appears to be fairly limited from this large scale efficient processing point of view. Out of very few examples, a linear-time algorithm for learning a scoring function and thresholding is presented in \cite{zhang2018tau}, which is still not an online algorithm (i.e. it is not designed to process data indefinitely on the fly) since batch processing is assumed with large space complexity and processing latency. Moreover, scoring of \cite{zhang2018tau} is similar to the one of \cite{zhao2009anomaly} but -unlike \cite{zhao2009anomaly}- trades off NP optimality for linear-time processing. Also, the technique of \cite{zhang2018tau} is restricted to linearly separable data only, and it requires to adjust thresholding for false positive rate controllability which can be seen impractical. The NP technique of \cite{ozkan2015online} is truly online (and one class) but it is strongly restricted to Markov sources, thus fails in the case of general non-Markov data (whereas our proposed algorithm has no such restriction). Another online NP classifier is presented in \cite{ONP} without strict assumptions unlike \cite{ozkan2015online}, but for only linearly separable data while leaving the online generalization to nonlinear setting as a future research direction.

To our best knowledge, online NP classification has not been studied yet in the nonlinear setting. Thus, as the first time in the literature, we solve the online and nonlinear NP classification problem based on a kernel inspired SLFN within the non-convex Lagrangian optimization framework of \cite{ONP,uzawa}, and use SGD updates for scalability. Our NP classifier exploits Fourier features \cite{Rahimi} and sinusoidal activations in the hidden layer of the SLFN (hence the name kernel inspired) to achieve a powerful nonlinear modeling with high computational efficiency and online real time processing capability.

Random Fourier features (RFFs) and also kernels in general have been successfully used for classification and regression of large scale data (please refer to \cite{Rahimi, lu2016large, porikli2011data} and \cite{wang2012breaking} for examples). Our presented work also exploits RFFs (during SLFN initialization) for large scale learning but, in contrast, for the completely different goal of solving the problem of online nonlinear Neyman-Pearson (NP) classification with neural networks in a non-convex Lagrangian optimization framework. Furthermore, the presented work learns the useful Fourier features with SGD updates beyond the initial randomness. On the other hand, kernels and RFFs have been previously studied in conjunction with neural networks. For example, computational relations from certain kernels to large networks are drawn in \cite{cho2009kernel}, and a kernel approximating convolutional neural network is proposed in \cite{mairal2014convolutional} for visual recognition. In particular, RFFs have been used to learn deep Gaussian processes \cite{cutajar2017random}, and for hybridization in deep models to connect linear layers nonlinearly \cite{mehrkanoon2018deep}. A radial basis function (rbf) network is proposed in \cite{other4} with batch processing, i.e., not online, which briefly discusses a heuristic by varying rbf parameters to manually control the false positive rate. Note that our SLFN is not an rbf network since we explicitly construct (during initialization) the kernel space in the hidden layer without a further need for kernel evaluations. We stress that the hidden layer of our SLFN for NP classification is same as the RFF layer of \cite{xie2019deep} for kernel learning (a simultaneous development of the same layer). The RFF layer in \cite{xie2019deep} is proposed as a building block to deep architectures for the goal of kernel learning. However, our goal of designing an online nonlinear NP classifier is completely different. Hence, our formulation, network objective and the resulting training process as well as our algorithm and experimental demonstration in this paper are fundamentally different compared to \cite{xie2019deep}. Moreover, online processing is not a focus in these studies except that \cite{mairal2014convolutional} and \cite{cutajar2017random} address scalability to voluminous data; and none of those (including \cite{xie2019deep} for kernel learning, and \cite{mairal2014convolutional} and \cite{cutajar2017random} for scalability) consider our goal of NP classification.


\section{Problem Description}\label{sec:Problem Desc}
Neyman-Pearson (NP) classification \cite{tong2016survey} seeks a classifier $\delta$ for a $d$ dimensional observation $\mathbb{R}^d\ni\vec{x}$ to choose one of the two classes $H_y: \vec{x} \sim p_y(\vec{x})$ as $\delta(\vec{x})=\hat{y}\in\{-1,+1\}$, where $y \in \{-1,+1\}$ (non-target: $-1$, target: $1$) is the true class label and $p_y(\vec{x})$ are the corresponding conditional probability density functions. The goal is to minimize the type II error (non-detection) rate $\text{P}_\text{nd}$
\begin{align}\label{eq:exp1}
\begin{split}
    \text{P}_\text{nd}(\delta)&=\int_{\forall \boldsymbol{x}\in \mathbb{R}^d} 1_{\{\hat{y}=-1\}} p_1(\vec{x}) d\vec{x}\\&=E_1[1_{\{\hat{y}=-1\}}]
\end{split}
\end{align}
(thus, the detection power $\text{P}_\text{td}=1-\text{P}_\text{nd}$ is maximized) while upper bounding the type I error $\text{P}_\text{fa}$ (false positive) rate by a user specified threshold $\tau$ as
\begin{align}\label{eq:exp2}
\begin{split}
    \text{P}_\text{fa}(\delta)&=\int_{\forall \boldsymbol{x}\in \mathbb{R}^d} 1_{\{\hat{y}=1\}} p_{-1}(\vec{x}) d\vec{x}\\&=E_{-1}[1_{\{\hat{y}=1\}}]\leq \tau
\end{split}
\end{align}
with $E_y$ being the corresponding expectations\footnote{In this paper, all vectors are column vectors and they are denoted by boldface lower case letters. For a vector $\vec{w}$, its transpose is represented by $\vec{w}'$ and the time index is given as subscript, i.e., $\vec{w}_t$. Also, a) $1_{\{\cdot\}}$ is the indicator function returning $1$ if its argument condition holds, and returning $0$, otherwise; and b) $\sgn(\cdot)$ is the sign function returning $1$ if its argument is positive, and returning $-1$, otherwise.}. Namely, $\delta^*$ is an NP classifier, if it satisfies
\begin{align*}
    \delta^*=\arg\min_{\delta} \text{P}_\text{nd}(\delta) \text{ subject to } \text{P}_\text{fa}(\delta) \leq \tau. 
\end{align*}
It is well-known that by the NP lemma \cite{scott2005neyman}, the likelihood ratio $\frac{p_{1}(\vec{x})}{p_{-1}(\vec{x})}$ provides an NP test, i.e.,
\begin{align}
\begin{split}
    &\delta^*(\vec{x})=-1, \text{ if } u(\vec{x})=\frac{p_{1}(\vec{x})}{p_{-1}(\vec{x})}-v(\tau) \leq 0, \text{ and }\\ &\delta^*(\vec{x})=1, \text{ otherwise, } \label{eq:NPdiscref}
\end{split}
\end{align}
where the offset $v(\tau)$ is chosen to satisfy the false positive rate constraint. Hence, finding the discriminant function $u$ is sufficient for NP testing.

The discriminant function $u$ can be simplified in many cases, and it might be linear or nonlinear as a function of $\vec{x}$ after full simplification. We provide two corresponding examples in the following. For instance, if the conditional densities $p_y(\vec{x})$ are both Gaussian with same covariances, then the discriminant is linear. On the other hand, in the example of one class classification \cite{scholkopf2001estimating} with applications to anomaly detection, there is typically no data from the target (anomaly) hypothesis because of the extreme rarity of anomalies, and there is also not much prior information due to the unpredictable nature of anomalies. Hence, the usual approach is to assume that the target density is uniform (with a finite support) \cite{zhao2009anomaly}, i.e., $p_1(\vec{x})=c$. Then, the critical region $\text{MVS}=\{\vec{x}\in \mathbb{R}^d:1/p_{-1}(\vec{x})\leq v(\tau)\}$
for the NP test to decide non-target, i.e., $\delta^*(\vec{x})=-1$, is known as the minimum volume set (MVS) \cite{scott2006learning} covering $1-\tau$ fraction of the non-target instances, i.e., $v(\tau)$ is set with simplification such that $\int_{\boldsymbol{x}\not\in \text{MVS}\subset \mathbb{R}^d}p_{-1}(\vec{x})d\vec{x}=\tau$.
Consequently, MVS has the minimum volume with respect to the uniform target density and hence maximizes the detection power. Here, the MVS discriminant $u(\vec{x})=1/p_{-1}(\vec{x})-v(\tau)$ (after simplification) is generally nonlinear, for instance, even when $p_{-1}(\vec{x})$ is Gaussian with zero mean unit-diagonal covariance. Therefore, we emphasize that the discriminant $u$ of the NP test\footnote{Note that knowing the continues valued discriminant $u$ is equivalent to knowing the discrete valued test $\delta^*$ due to one-to-one correspondence, i.e., $\delta^*(\boldsymbol{x})=\sgn(u(\boldsymbol{x}))$. Hence, in the rest of the paper, we refer to the discriminant as the NP classifier as well.} might be arbitrarily nonlinear in general. Furthermore, since the discriminant definition requires the knowledge of the conditional densities $p_y$ which are unavailable in most realistic scenarios, the discriminant $u$ is unknown. For this reason, NP classification refers to the data driven statistical learning of an approximation $f^*\in\mathcal{H}$ of the unknown discriminant $u$ based on given two classes of data $\{(\vec{x}_t,y_t)\}$, where $\mathcal{H}$ is an appropriate set of functions which is sufficiently powerful to model the complexity of $u$.

As a result, the data driven statistical learning of the NP classifier $f^*$ is obtained as the output of the following NP optimization:
\begin{align}
    u\simeq f^*=\arg\min_{f\in\mathcal{H}} \hat{\text{P}}_\text{nd}(f) \text{ subject to } \hat{\text{P}}_\text{fa}(f) \leq \tau, \label{eq:NP_general_formulation}
\end{align}
where
\begin{align*}
    &\hat{\text{P}}_\text{nd}(f)=\frac{\sum_{\forall t: y_t=1}1_{\{f(\boldsymbol{x}_t)\leq0\}}}{\sum_{\forall t: y_t=1}1} \text{ and }\\ &\hat{\text{P}}_\text{fa}(f)=\frac{\sum_{\forall t: y_t=-1}1_{\{f(\boldsymbol{x}_t)>0\}}}{\sum_{\forall t: y_t=-1}1}
\end{align*}
empirically estimates the type I (expectation in \eqref{eq:exp1}) and type II (expectation in \eqref{eq:exp2}) errors, respectively. For example, \cite{ONP} studies this optimization in \eqref{eq:NP_general_formulation} for the set $\mathcal{H}$ of linear discriminants, in which case -however- the resulting linear NP classifier is largely suboptimal in most realistic scenarios; for example, the MVS estimation for anomaly detection requires to learn nonlinear class separation boundaries with a nonlinear discriminant.

\begin{figure}[t!]
	\centering
	\includegraphics[width=.75\linewidth]{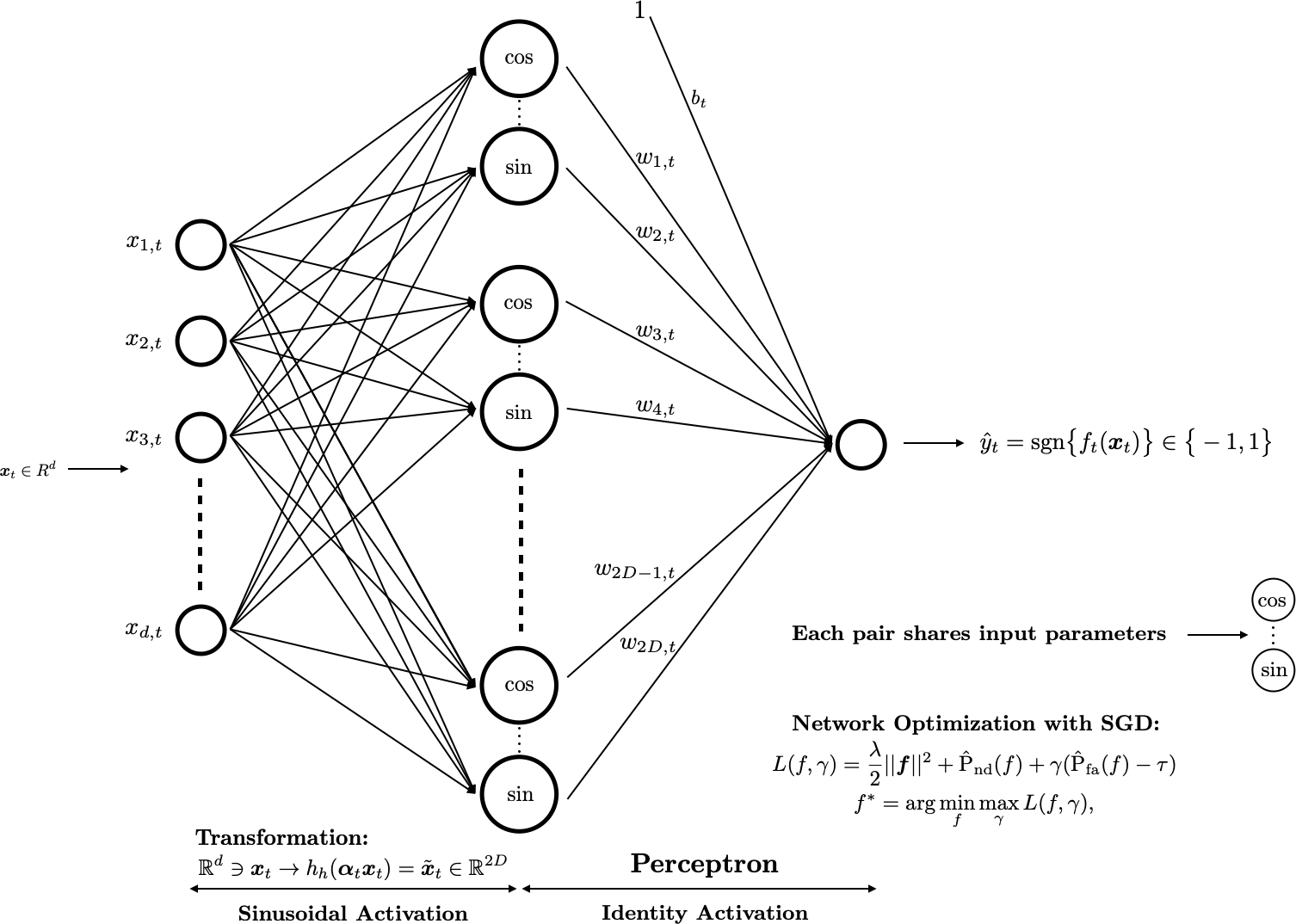}
	\caption{The single hidden layer feed forward neural network (SLFN) that we use for online nonlinear Neyman-Pearson (NP) classification is illustrated. The hidden layer is initialized to approximately construct the high dimensional kernel space (e.g., radial basis function) via random Fourier features (RFFs) with sinusoidal activation. The output layer follows with identity activation. This network is compact and strongly nonlinear with expedited learning ability thanks to i) the exponential convergence of the inner products (w.r.t. the number of hidden nodes) in the space of RFFs to the true kernel with an excellent random network initialization, and ii) the learning of Fourier features (instead of relying on randomization) by the data driven network updates. We learn the network parameters sequentially via SGD based on a nonconvex Lagrangian NP objective. The result is an exptedited powerful nonlinear NP modeling with high computational efficiency and scalibility.}
	\label{fig:elm}
\end{figure}

Our goal in the presented work is to develop, as the first time in the literature to our best knowledge, an online nonlinear NP classifier for any given user-specified desired false positive rate $\tau$ with real time processing capability. In particular, we use a kernel inspired single hidden layer feed forward neural network (SLFN), cf. Fig. \ref{fig:elm}, to model the set $\mathcal{H}$ of nonlinear candidate discriminant functions in \eqref{eq:NP_general_formulation} as
\begin{align}
    &\mathcal{H}=\{f:f(\boldsymbol{x})=h_o(h_h(\boldsymbol{\alpha}\boldsymbol{x})'\boldsymbol{w}+b),\forall \boldsymbol{\alpha}, \forall \boldsymbol{w}, \forall b\}, \label{network_space}
\end{align}
where $\boldsymbol{\alpha}$ and $(\boldsymbol{w},b)$ are the hidden and output layer parameters, and $h_h$ and $h_o$ are the nonlinear hidden and identity output layer activations. We sequentially learn the SLFN parameters based on the NP objective (that is maximizing the detection power about a user-specified false positive rate as given in \eqref{eq:NP_general_formulation}) with stochastic gradient descent (SGD) to obtain the nonlinear classification boundary, i.e., to estimate the unknown discriminant $u$, in an online manner while maintaining scalibility to voluminous data.

The data processing in our proposed algorithm is computationally highly efficient and truly online with $O(N)$ computational and $O(1)$ space complexity ($N$ is the total number of processed instances). Namely, we sequentially observe the data $\vec{x}_t \in \mathbb{R}^d$ indefinitely without knowing a horizon, and decide about its label $\hat{y}_t \in \{1,-1\}$ as $\hat{y}_t=1$ if the SLFN ${f}_{t}\in\mathcal{H}$ at time $t$ provides ${f}_{t}(\vec{x}_{t})> 0$, and as $\hat{y}_t=-1$, otherwise. Then, we update our model ${f}_{t}$, i.e., update the SLFN at time $t$, to obtain ${f}_{t+1}\in\mathcal{H}$ based on the error $y_t-\hat{y}_t$ via SGD and discard the observed data, i.e., $\vec{x}_t$ and $y_t$, without storing. Hence, each instance is processed only once. In this processing framework, $f_t\rightarrow f^*\in \mathcal{H}$ models the NP discriminant $u$ in \eqref{eq:NPdiscref}. As a result of this processing efficiency, our algorithm is appropriate for large scale data applications.

\section{SLFN for Online Nonlinear NP Classification}\label{sec:Method}
In order to learn nonlinear Neyman-Pearson classification boundaries, we use a single hidden layer feed forward neural network (SLFN), illustrated in \ref{fig:elm}, that is designed based on the kernel approach to nonlinear modeling (cf. \cite{Rahimi} and the references therein for the mentioned kernel approach). Namely, the hidden layer is randomly initialized to explicitly transform the observation space (via $\phi_{\vec{\alpha_1}}$) into a high dimensional kernel space with sinusoidal hidden layer activations by using the random Fourier features \cite{Rahimi}. We use a certain variant of the perceptron algorithm \cite{rosenblatt1958perceptron} as the output layer with identity activation followed by a sigmoid loss. Based on this SLFN, we sequentially (in a truly online manner) learn the network parameters, i.e., the classifier parameters $\vec{w}_t, b_t$ as well as the kernel mapping parameters $\vec{\alpha_t}$, through SGD in accordance with the NP optimization objective \eqref{eq:NP_general_formulation}.

{\bf In the hidden layer} of the SLFN, the randomized initial transformation $\phi_{\vec{\alpha_1}}
: \mathbb{R}^d \rightarrow \mathbb{R}^{2D}$ at time $t=1$, 
\begin{align}
\mathbb{R}^d \ni \vec{x} \rightarrow \tilde{\vec{x}} = \phi_{\vec{\alpha_1}}(\vec{x}) \in \mathbb{R}^{2D}, \label{eq:phi}
\end{align}
is constructed based on the fact (as provided in \cite{Rahimi}) that any continuous, symmetric and shift invariant kernel can be approximated as $k(\vec{x}^i, \vec{x}^j) \triangleq  k(\vec{x}^i-\vec{x}^j) \approx \phi_{\vec{\alpha_1}}({\vec{x}^i})' \phi_{\vec{\alpha_1}}({\vec{x}^j})$
with an appropriately randomized kernel feature mapping. Note that the kernel $k(\vec{x}^i, \vec{x}^j)$ is an implicit access to the targeted high dimensional kernel space as it encodes the targeted inner products. This kernel space is explicitly and approximately constructed by the sinusoidal hidden layer activations of the SLFN in which the new inner products across activations approximate originally targeted inner products. Hence, linear techniques applied to the sinusoidal hidden layer activations can learn nonlinear models. In our method, we use the radial basis function (rbf) kernel\footnote{We use the rbf kernel in this study as an example but it is not required. Thus, the presented technique can be straightforwardly extended to any symmetric and shift invariant kernel satisfying the Bochner's theorem, cf. \cite{Rahimi}.} $k(\vec{x}^i, \vec{x}^j) = \exp(-g ||\vec{x}^i-\vec{x}^j||^2)$ with the bandwidth parameter $g$ (that is inversely related to the actual bandwidth). Then, 
\begin{align}
    k(\vec{x}^i-\vec{x}^j) = E_{\vec{\bar{\alpha}_1}} [r_{\vec{\bar{\alpha}_1}} (\vec{x}^i) r_{\vec{\bar{\alpha}_1}} (\vec{x}^j)'] \label{eq:this_exp}
\end{align}
(due to Bochner's theorem as provided in \cite{Rahimi}) where the Fourier feature is 
\begin{align}
    r_{\vec{\bar{\alpha}_1}} (\vec{x})=[\cos(\vec{\bar{\alpha}_1}'\vec{x}), \sin(\vec{\bar{\alpha}_1}'\vec{x})] \label{eq:FF}
\end{align}
and $\vec{\bar{\alpha}_1}$ is sampled from the $d$ dimensional multivariate Gaussian distribution $p(\vec{\bar{\alpha}_1})=N(\vec{0}, 2g\vec{I})$ (which is the Fourier transform of the kernel in hand) with $E_{\vec{\bar{\alpha}_1}}$ being the corresponding expectation. Hence, by replacing the expectation in \eqref{eq:this_exp} with the independent and identically distributed (i.i.d) sample mean of the ensemble $\{r_{\vec{\bar{\alpha}^q_1}} (\vec{x}^i) r_{\vec{\bar{\alpha}^q_1}} (\vec{x}^j)'\}_{q=1}^{D}$ 
of size $D$, we define our kernel mapping as
\begin{align}\label{mapping}
\begin{split}
\tilde{\vec{x}}=&\phi_{\vec{\alpha_1}}(\vec{x})\\=&\sqrt{\frac{1}{D}}[r_{\vec{\bar{\alpha}_1^1}}(\vec{x}),r_{\vec{\bar{\alpha}_1^2}}(\vec{x}),\cdots,r_{\vec{\bar{\alpha}_1^D}}(\vec{x})]',
\end{split}
\end{align}
\noindent
which can be directly implemented in the hidden layer of the SLFN, cf. Fig. \ref{fig:elm}, along with the sinusoidal activation due to the definition of $r_{\vec{\bar{\alpha}_1}}$.

Note that  $\vec{\alpha_t}$ keeps all the hidden layer parameters at time $t$ as a matrix of size $2D\times d$ consisting of $\vec{\bar{\alpha}^i_t}$'s corresponding to the hidden units, i.e., $\vec{\alpha_t}=[\vec{\bar{\alpha}^1_t},\vec{\bar{\alpha}^1_t},\vec{\bar{\alpha}^2_t},\vec{\bar{\alpha}^2_t},\cdots,\vec{\bar{\alpha}^D_t},\vec{\bar{\alpha}^D_t}]'$. And the hidden layer activation is sinusoidal: $h_h(m)=\cos(m)$ and $h_h(m)=\sin(m)$ for the odd and even indexed hidden nodes, respectively, due to the definition in \eqref{eq:FF}. At time $t=1$, $\vec{\alpha_1}$ is randomly initialized with an appropriate $g$ of the rbf kernel so that the SLFN starts with approximately constructing the high dimensional kernel space $\bar{\mathcal{H}}=\{f:f(\boldsymbol{x})=h_o(h_h(\boldsymbol{\alpha}_1\boldsymbol{x})'\boldsymbol{w} +b), \forall \boldsymbol{w}, \forall b\}$ in its hidden layer, and in relation to \eqref{eq:phi}, $\tilde{\vec{x}}=\phi_{\vec{\alpha_1}}(\vec{x})=h_h(\boldsymbol{\alpha}_1\boldsymbol{x})$. Note that $\bar{\mathcal{H}}$ of the rbf kernel readily provides a powerful nonlinear modeling to the SLFN even if the hidden layer is kept untrained. Thanks to this excellent network initialization, we achieve an expedited process of learning from data. Moreover, in the course of our sequential processing, the SLFN continuously updates and improves the hidden layer, i.e., kernel mapping, parameters as $\vec{\alpha_t}$. Therefore, we optimize a nonlinear NP classifier in actually the larger space $\mathcal{H}\supset \bar{\mathcal{H}}$ (as our optimization is not restricted to $\vec{\alpha_1}$ of the random initialization, cf. the definition of $\mathcal{H}$ in \eqref{network_space}) for greater nonlinear modeling capability compared to the rbf kernel.

The SLFN in Fig. \ref{fig:elm} that we use for online and nonlinear NP classification is compact in principle since the required number of hidden nodes is relatively small. The reason is that the convergence of the sample mean of the i.i.d. ensemble $\{r_{\vec{\bar{\alpha}^q_1}} (\vec{x}^i) r_{\vec{\bar{\alpha}^q_1}} (\vec{x}^j)'\}_{q=1}^{D}$ of size $D$ to the true mean $k(\vec{x}^i, \vec{x}^j)$ is exponentially fast with the order of $O(e^{-D})$ by Hoeffding's inequality \cite{Rahimi}. On the other hand, since random Fourier features are independent of data, further compactification is possible by eliminating irrelevant, i.e., unuseful, Fourier features in a data driven manner, cf. the examples of feature selection in \cite{porikli2011data} and Nystr{\"o}m method in \cite{yang2012nystrom} for this purpose. In contrast, and alternatively, we distill useful Fourier features in the hidden layer activations as a result of the sequential learning of the kernel mapping parameters, i.e., $\vec{\bar{\alpha}^i_t}$, via SGD. Hence, nodes of the SLFN are dedicated to only useful Fourier features, and thus we achieve a further network compactification by reducing the necessary number of hidden nodes as well as reducing the parameter complexity. Then, one can expect to better fight overfitting with great nonlinear modeling power and NP classification performance. This compactification does also significantly reduce the computational as well as space complexity of our SLFN based classifier, which -together with the SGD optimization- yields scalability to voluminous data. Consequently, the proposed online NP classifier is computationally highly efficient and appropriate for real time processing in large scale data applications.

{\bf Remark : }We obtain a sequence of kernel mapping parameters $\vec{\alpha}_t$ in the course of data processing. This means that at the end of processing $N$ instances, one can potentially construct a new non-isotropic rbf kernel by estimating the multivariate density of the collection $\{\vec{\bar{\alpha}}^j_N\}_{j=1}^{D}$ (here, we assume that $D$ is large and the density is multivariate Gaussian. If it is not Gaussian, then one can straightforwardly incorporate a Gaussianity measure into the overall network objective) and then finding out the corresponding non-isotropic rbf kernel by taking back the inverse Fourier transform of the estimated density. Therefore, our algorithm is also kernel-adaptive since it essentially learns a new kernel (and also improves the previous one) at each SGD learning step. This kernel adaptation ability can be improved. For instance, one can start with a random mapping as described and estimate the density of the mapping parameters after convergence, and then re-start with new samples from the converged density. Multiple iterations of this process may yield better kernel adaptation (but re-running would hinder online processing and define batch processing, hence it is out of scope of the present work), which we consider as future work.

{\bf In the output layer} of the SLFN, we use a certain variant of perceptron \cite{rosenblatt1958perceptron} with the identity activation, i.e., $h_o(m)=m$. Then, the classification model is defined linearly after the hidden layer kernel inspired transformation  as $f(\vec{x})= h_o(\langle\vec{w},\tilde{\vec{x}} \rangle) +b= \langle\vec{w},\tilde{\vec{x}} \rangle +b=h_h(\boldsymbol{\alpha}\vec{x})'\vec{w} +b$,
where $\vec{w} \in \mathbb{R}^{2D}$ is the normal vector to the linear separator and $b \in \mathbb{R}$ is the bias. Thus, the decision of the SLFN is $\hat{y}=\sgn(f(\vec{x}))$.

{\bf Regarding the overall network objective} for sequential learning of the network parameters $\boldsymbol{\alpha}_t,\boldsymbol{w}_t,b_t$ and solving the NP optimization in \eqref{eq:NP_general_formulation} to obtain our SLFN based online nonlinear NP classifier, we next formulate the NP objective similar to \cite{ONP} as
\begin{align}\label{NPproblem}
\begin{split}
f^*=&\arg\min_{f\in\mathcal{H}} {}\frac{\lambda}{2}||f||^2+\hat{\text{P}}_\text{nd}(f)\\&\text{  subject to  } \hat{\text{P}}_\text{fa}(f) \leq\tau,
\end{split}
\end{align}
where the first term $\frac{\lambda}{2}||f||^2$ is the regularizer for which we use the magnitude of the classifier parameters in the output layer, i.e., $\frac{\lambda}{2}||\boldsymbol{w}||^2$, and $\lambda$ is the regularization weight. For differentiability,  the non-detection $\hat{\text{P}}_\text{nd}$ and false positive $\hat{\text{P}}_\text{fa}$ error rates are estimated based on data until time $t$ as
\begin{align}\label{Pnd}
\begin{split}
&\hat{\text{P}}_\text{nd}(f)=\frac{1}{n_{t_{+}}} \sum_{t'\in S_{1}^t}^{}l(f(\vec{x}_{t'})) \text{ and }\\&\hat{\text{P}}_\text{fa}(f)=\frac{1}{n_{t_{-}}} \sum_{S_{-1}^t}^{}l(-f(\vec{x}_{t'}))
\end{split}
\end{align}
with $S_c^t=\{t':1\leq {t'}\leq t, y_{{t'}}=c\}$, $n_{t_{+}}=|S_1^t|$ (set cardinality) and $n_{t_{-}}=|S_{-1}^t|$. Note that another appropriate function can be used here to obtain a differentiable surrogate for the $0-1$ errors in \eqref{eq:NP_general_formulation} for estimating the error rates. However, our results in the rest of this paper are based on the sigmoid loss $l(m)=\frac{1}{1+\exp (m)}$.

For sequential optimization of the NP objective in \eqref{NPproblem}, we next define the following Lagrangian
\begin{align}
L(f, \gamma) &= \frac{\lambda}{2}||\vec{f}||^2+\hat{\text{P}}_\text{nd}(f) + \gamma(\hat{\text{P}}_\text{fa}(f)- \tau) \label{lagrangian},
\end{align}

\noindent
where $\tau$ is the user-specified desired false positive rate and $\gamma \in \mathbb{R}^+$ is the corresponding Lagrange multiplier.

Since the saddle points of \eqref{lagrangian} correspond to the local minimum of \eqref{NPproblem}, cf. \cite{ONP} and \cite{uzawa} for the details, we apply the Uzawa approach \cite{uzawa} to search for the saddle points of \eqref{lagrangian} and learn our parameters in the online setting with SGD updates. To be more precise, we follow the optimization framework of \cite{ONP} and solve the $\min\max$ optimization $f^*=\arg\min_f\max_\gamma L(f, \gamma)$ via an iterative approach with gradient steps, where one iteration minimizes $L(f, \gamma)$ for a fixed $\gamma$ and the other maximizes $L(f, \gamma)$ for a fixed $f$. Note that the fixed-$\gamma$ minimization

\begin{align*}
    \arg\min_{f\in\mathcal{H}}L(f, \gamma)=\arg\min_{f\in\mathcal{H}}&\frac{\lambda}{2}||\vec{f}||^2 + \hat{\text{P}}_\text{nd}(f) + \gamma(\hat{\text{P}}_\text{fa}(f)- \tau)\\
    =\arg\min_{f\in\mathcal{H}}&\frac{\lambda}{2}||\vec{f}||^2+\hat{\text{P}}_\text{nd}(f) + \gamma\hat{\text{P}}_\text{fa}(f)
\end{align*}
is a regularized weighted error minimization, where the ratio of the type I error rate cost to the one of type II error rate is $\gamma$. Hence, the unknown Lagrange multiplier $\gamma$ defines (up to a scaling with the prior probabilities) the asymmetrical error costs that correspond to the false positive rate constraint in \eqref{eq:NP_general_formulation}. On the other hand, the gradient ascent updates $\gamma \leftarrow \gamma + \beta\nabla_\gamma L(f,\gamma)=\gamma +\beta(\hat{\text{P}}_\text{fa}(f)- \tau)$ in the fixed-$f$ maximization determines the unknown multiplier $\gamma$ so that the type I error cost is decreased (increased) if the error estimate is below (above) the tolerable rate $\tau$ in favor of detection power (true negative detection). This provides an iterative learning of the correspondence between the asymmetrical error costs and the NP constraint.

To this end, inserting the definitions in \eqref{Pnd} and \eqref{Pnd} into \eqref{lagrangian} with the regularization $\frac{\lambda}{2}||f||^2= \frac{\lambda}{2}||\boldsymbol{w}||^2$ yields the overall SLFN objective as follows
\begin{align}\label{l2}
\begin{split}
L(f, \gamma) =& \frac{\lambda}{2}||\boldsymbol{w}||^2+\frac{1}{n_{t_{+}}} \sum_{1\leq {t'}\leq t: y_{{t'}}=1}^{}l\Big(y_{t'} f(\vec{x}_{t'})\Big) 
\\
&+ \frac{\gamma}{n_{t_{-}}} \sum_{1\leq {t'}\leq t: y_{{t'}}=-1}^{}l\Big(y_{{t'}} f(\vec{x}_{t'})\Big)- \gamma\tau
\\
=&\frac{1}{t}\sum_{{t'}=1}^{t} \bigg(\frac{\lambda}{2}||\vec{w}||^2 +\mu_{t'} l\Big(y_{t'} f(\vec{x}_{t'})\Big) -\gamma \tau \bigg) 
\\
=&\frac{1}{t}\sum_{{t'}=1}^{t} s(f,\gamma,t'),
\end{split}
\end{align}
where $s(f,\gamma,t')=\bigg(\frac{\lambda}{2}||\vec{w}||^2 +\mu_{t'} l\Big(y_{t'} f(\vec{x}_{t'})\Big) -\gamma \tau \bigg)$ and $\mu_{t'}=\frac{t}{n_{t_{+}}}$ if $y_{t'}=+1$, and $\gamma\frac{t}{n_{t_{-}}}$, otherwise.

In order to learn the SLFN parameters for obtaining the proposed online nonlinear NP classifier via the NP optimization explained above, we use stochastic gradient descent (SGD) to sequentially optimize the overall network objective defined in \eqref{l2}. These network parameters are 1) $\boldsymbol{\alpha}$, to project input $\boldsymbol{x}$ to the higher dimensional kernel space, 2) $\boldsymbol{w}$ and $b$, which are the perceptron parameters of the output layer to classify the projected input $\tilde{\vec{x}}$, and 3) $\gamma$, to learn the correspondence between the error costs and the NP constraint.

Suppose at the beginning of time $t$, we have an existing model $f_{t}$ learned with the past data as well as the error costs corresponding to $\gamma_{t}$; and a little later, we observe the instance $\boldsymbol{x}_t$. SGD based optimization takes steps to update $f_t$ and $\gamma_t$ to obtain $f_{t+1}$ and $\gamma_{t+1}$ with respect to the partial derivatives of the instantaneous objective $s(f_t,\gamma_t,t)$. Namely, $f_{t+1}=f_t-\eta_t\nabla_f s(f_t,\gamma_t,t) \text{ and }\gamma_{t+1}=\gamma_t+\beta_t\nabla_\gamma s(f_t,\gamma_t,t).$ Based on the partial derivatives of the instantaneous objective $s(f_t,\gamma_t,t)$ defined in \eqref{l2}, the SGD updates for the SLFN parameters can be computed $\forall i \in\{1,\cdots,D\}$ as $\vec{w}_{t+1}=\vec{w}_{t} - \eta_t \Big(\lambda \vec{w}_{t} + \mu_{t} \nabla_{\vec{w}}l\big(y_{t} f_t(\vec{x}_t)\big)\Big)$, $b_{t+1}= b_{t} -\eta_t \Big(\mu_{t} \nabla_{b}l\big(y_{t} f_t(\vec{x}_t)\big)\Big)$, $\vec{\bar{\alpha}^i_{t+1}}=\vec{\bar{\alpha}^i_t}-\eta_t\Big(\mu_t \nabla_{\vec{\bar{\alpha}^i}}l\big(y_{t} f_t(\vec{x}_t)\big) \Big)$, and $\gamma_{t+1}=\gamma_{t} +\beta_t \Big((1_{\{y_t=-1\}}\frac{t}{n_{t_-}})l\big(y_{t} f_t(\vec{x}_t)\big)-\tau \Big)$, where $\eta_t$ is the learning rate and $\beta_t$ is named as the Uzawa gain \cite{uzawa} controlling the learning rate of the Lagrange multiplier. Using the sigmoid $l(m)=\frac{1}{1+\exp (m)}$ yields the partial derivatives with $\tilde{\vec{x}}_t= h_h(\boldsymbol{\alpha}_t\vec{x}_t)$ as
\begin{align}
    &\nabla_{\vec{w}}l\big(y_{t} f_t(\vec{x}_t))\big) = -\tilde{\vec{x}}_tl^2(y_{t} f_t(\vec{x}_t))\exp(y_{t} f_t(\vec{x}_t))y_t, \label{eq:u1}\\
    &\nabla_{b}l\big(y_{t} f_t(\vec{x}_t))\big) = -l^2(y_{t} f_t(\vec{x}_t))\exp(y_{t} f_t(\vec{x}_t))y_t, \text{ and } \label{eq:u2}\\
&\nabla_{\vec{\bar{\alpha}^i}}l\big(y_{t} f_t(\vec{x}_t))\big)= -\vec{x}_tl^2(y_{t} f_t(\vec{x}_t))\exp(y_{t} f_t(\vec{x}_t))y_t \label{eq:u3}\\
 &\times \big(-{w}^{2i-1}_t \sin(\vec{\bar{\alpha}^{i'}_t}\vec{x}_t) +{w}^{2i}_t \cos(\vec{\bar{\alpha}^{i'}_t}\vec{x}_t)  \big), \nonumber
\end{align}
which can be straightforwardly incorporated into the backpropagation.

In our experiments, we obtain an empirical false positive rate estimate $\hat{\text{P}}_\text{fa}$ based on a sliding window keeping the $0-1$ errors for a couple hundreds of the past negative data instances, and use the following $\gamma$ update instead of the aforementioned stochastic one: 
\begin{align}
\label{gamma}
\gamma_{t+1}=\gamma_{t}\bigg(1+\beta_t \Big(\hat{\text{P}}_\text{fa}-\tau \Big)\bigg), \text{ when $y_t=-1$},
\end{align}
which has been observed to yield a more stable and robust performance. Note that this update is directly resulted from \eqref{lagrangian}, and does certainly not disturb real-time online processing since a past window of positive decisions requires almost no additional space complexity (only $200$ bits in the case of, for instance, storing binary decisions for $200$ past negative instances).

\begin{algorithm}[t!]
	\caption{Proposed Online Nonlinear Neyman-Pearson Classifier (NP-NN)}\label{alg:Theorem1}
	\begin{algorithmic}[1]
		\STATE Set the desired (or target) false positive rate (TFPR) $\tau$, regularization $\lambda$, number $2D$ of hidden nodes and bandwidth $g$ for the rbf kernel
		\STATE Initialize the SLFN parameters $\vec{\alpha}_1$, $\vec{w}_1$, $b_1$, and learning rates $\eta_1$, $\beta_1$, $\gamma_1$ 
		\STATE Set $n_{t_+}=n_{t_-}=0$, and sliding window size $W_s=200$
		\FOR{$t=1,2,\dots$}
		\STATE Receive $\vec{x}_t$ and calculate $\tilde{\vec{x}}_t= h_h(\boldsymbol{\alpha}_t\vec{x}_t)$ and $f_t(\vec{x}_t)=\vec{w}_t'\tilde{\vec{x}}_t+b_t$
		\STATE Calculate the current decision as $\hat{y}_t=\sgn(f_t(\vec{x}_t))$ and observe $y_t$
		\STATE Calculate $n_{t_+}=n_{t_+}+1_{\{y_t=1\}}$ and $n_{t_-}=n_{t_-}+1_{\{y_t=-1\}}$ 
		\STATE Calculate $\mu_t=\frac{t}{n_{t_+}}1_{\{y_t=1\}}+\frac{\gamma t}{n_{t_-}}1_{\{y_t=-1\}}$
		\STATE Update $\vec{w}_{t+1}=\vec{w}_{t} - \eta_t \Big(\lambda \vec{w}_{t} + \mu_{t} \nabla_{\vec{w}}l\big(y_{t} f_t(\vec{x}_t)\big)\Big)$, cf. \eqref{eq:u1}
		\STATE Update $b_{t+1}= b_{t} -\eta_t \Big(\mu_{t} \nabla_{b}l\big(y_{t} f_t(\vec{x}_t)\big)\Big)$, cf. \eqref{eq:u2}
		\STATE Update $\vec{\bar{\alpha}^i_{t+1}}=\vec{\bar{\alpha}^i_t}-\eta_t\Big(\mu_t \nabla_{\vec{\bar{\alpha}^i}}l\big(y_{t} f_t(\vec{x}_t)\big) \Big)$, cf. \eqref{eq:u3}
		\STATE Update $\gamma_{t+1}=\gamma_{t}\bigg(1+\beta_t \Big(\hat{\text{P}}_\text{fa}-\tau \Big)\bigg)$, if $y_t=-1$, cf. \eqref{gamma} and the explanation about the estimate $\hat{\text{P}}_\text{fa}$ of the false positive rate
		\STATE Update $\eta_{t+1}=\eta_{1}(1+\lambda t)^{-1} $ and $\beta_{t+1}=\beta_{1}(1+\lambda t)^{-1}$
		\ENDFOR
	\end{algorithmic}
\end{algorithm}

Based on the derivations above, we sequentially update the SLFN at each time in a truly online manner with $O(N)$ (here, $N$: total number of processed data instances) computational and $O(1)$ space complexity in accordance with the NP objective. Hence, we construct our method called ``NP-NN" in Algorithm \ref{alg:Theorem1} that can be used in real time for online nonlinear Neyman-Pearson classification. We refer to the Section \ref{sec:Experiments} of our experimental study for all the details about the input parameters and initializations.

{\bf Remark: }Recall that the goal in NP classification is to achieve the minimum miss rate (maximum detection power) while upper bounding the false positive rate (FPR) by a user-specified threshold $\tau$. Therefore, both aspects (minimum miss rate and its FPR constraint) of this goal should be considered in evaluating the performance of NP classifiers. The NP-score of \cite{scott2007performance,davenport2010tuning} is defined as
\begin{align}
    \text{NP-score} = \kappa \max(\hat{P}_{\text{fa}}(f)-\tau,0)+\hat{P}_{\text{nd}}(f),
    \label{eq:NP_score_kappa}
\end{align}
where $f$ is the NP model to be evaluated and $\kappa$ controls the relative weights of the miss rate (with weight $1$) and its FPR constraint (with weight $\kappa$ if the desired rate is exceeded, and with weight $0$ otherwise). Namely, $\kappa$ controls the hardness of the NP FPR constraint, and a smaller NP-score indicates a better NP classifier. By enforcing a strict hard constraint on FPR with a very large $\kappa \simeq \infty$, one can immediately reject models (while evaluating various models) that violate FPR constraint with even a slight positive deviation from the desired FPR $\tau$ (a negative deviation does not violate). However, even though the original NP formulation requires a hard constraint, we consider that it is not appropriate to use a hard constraint in practice, as also extensively explained in \cite{scott2007performance}, based on the following two reasons: (1) An NP classifier is typically learned using a set of observations, and that set is itself a random sample from the underlying density of the data. Hence, the estimated FPR $\hat{P}_{\text{fa}}(f)$ of the model is also a random quantity, which is merely an estimator of the unknown true FPR $P_{\text{fa}}(f)$. Note that the true FPR $P_{\text{fa}}(f)$ is actually the one to be strictly constrained, but unavailable. Thus, it is unreliable to enforce a strict hard constraint (with a very large $\kappa\simeq \infty$) on the random estimator $\hat{P}_{\text{fa}}(f)$, and a relatively soft constraint has surely more practical value by allowing a small positive deviation from the desired FPR $\tau$. (2) Also, one might be willing to exchange true negatives in favor of detections with a small positive deviation from the desired FPR $\tau$, when the gain is larger than the loss as the NP-score improves. Consequently, for parameter selections with cross validation in our algorithm design as well as for performance evaluations in our experiments, we opt for a relatively soft constraint and use $\kappa=1/\tau$ in accordance with the recommendation by the authors \cite{scott2007performance}. This choice allows a relatively small positive deviation from the desired FPR, and normalizes the deviation by measuring it in a relative percentage manner. For example, the positive deviations $0.1$ and $0.001$ both degrade the score equally by $50\%$ when the desired rates are $0.2$ and $0.002$, respectively. Various other NP studies in the literature do also practically allow small positive deviations from the desired FPR $\tau$. For instance, we observe such a deviation in \cite{ONP} with theirs and compared algorithms \cite{davenport2010tuning} in the case of spambase dataset, in \cite{zhang2018tau} with theirs in the case of heart and breast cancer datasets, and finally in \cite{kong2019false} with one of the compared algorithms \cite{du2015convex} in all datasets.

A comprehensive experimental evaluation of our proposed technique is next provided based on real as well as synthetic datasets in comparison to state-of-the-art competing methods.

\section{Experiments}\label{sec:Experiments}
We present extensive comparisons of the proposed kernel inspired SLFN for online nonlinear Neyman-Pearson classification (NP-NN), described in Algorithm \ref{alg:Theorem1}, with $3$ different state-of-the-art NP classifiers. These compared techniques are online linear NP (OLNP) \cite{ONP}, as well as logistic regression (NPROC-LOG) \cite{cox1958regression} and support vector machines with rbf kernel (NPROC-SVM) \cite{cortes1995support} in the NP framework of the umbrella algorithm described in \cite{tong2018neyman}. Among these, OLNP (linear NP classification) is an online technique with $O(N)$ computational complexity, whereas NPROC-LOG (linear NP classification) and NPROC-SVM (nonlinear NP classification) are batch techniques with at least $O(N^2)$ computational complexity, where $N$ is the number of processed instances. In contrast, we emphasize that to our best knowledge, the proposed NP classifier NP-NN is both nonlinear and online as the first time in literature, with $O(N)$ computational and negligible space complexity  resulting real time nonlinear NP modeling and false positive rate controllability. Consequently, the proposed NP-NN is appropriate for challenging fast streaming data applications.

\begin{figure*}[t!]
    \centering
        \includegraphics[width=\linewidth]{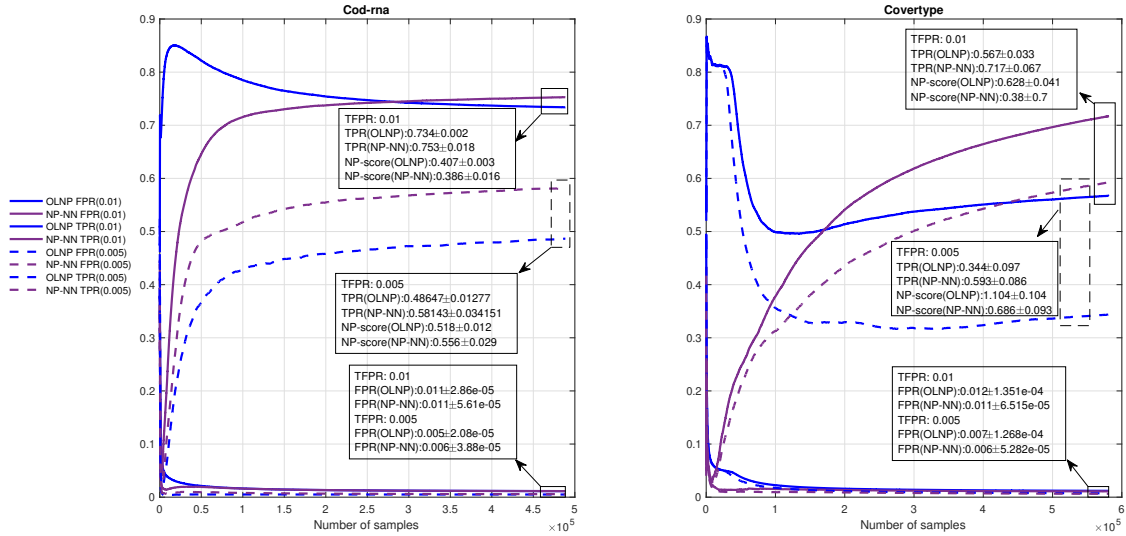}
    \caption{We demonstrate the proposed online algorithm NP-NN on two large scale datasets (Cod-rna and Covertype) with relatively small target false positive rates, i.e., $\text{TFPR}\in \{0.01, 0.005\}$. The data processing in this case is truly online, and based on only a single pass over the stream without separate training and test phases. Hence, this experiment better demonstres the typical use-case of our algorithm in large scale scenarios. Time accumulated false positive error rate (FPR) and detection rate (TPR) are obtained after averaging over $15$ trials of random data permutations. Overall, we observe that the proposed NP-NN and OLNP are both decent and comparable in terms of the false positive rate controllability, whereas the proposed NP-NN strongly outperforms OLNP in terms of both the detection power and NP-score.}
    \label{fig:large_dataset_conv}
\end{figure*}

We conduct experiments based on various real and synthetic datasets \cite{Dua:2019, CC01a} from several fields such as bioinformatics and computer vision, each of which is normalized by either unit-norm (each instance is divided by its magnitude) or z-score (each feature is brought down to zero mean unit variance) normalization before processing. For each dataset, smaller class is designated as the positive (target) class. The details of the datasets are provided in Table \ref{tab:NP_scores_and_AUC}, where the starred ones and unstarred ones are normalized with unit norm and z-score, respectively. For performance evaluations, we generate $15$ random permutations of each dataset, and each random permutation is split into two as training ($\% 75$) and test ($\% 25$) sequences. We strongly emphasize that the processing in the proposed algorithm NP-NN is truly online, meaning that, there are no separate training and test phases. However, since NPROC-LOG and NPROC-SVM are batch algorithms requiring a separate training, we opt to use training/test splits in this first set of experiments for a fair and statistically unbiased robust performance comparison. Such a split is in fact not needed in practice in the case of the proposed NP-NN that -by design- processes data on the fly. Additional experiments based on two larger scale datasets to demonstrate the ideal use-case (i.e. online processing without separate training/tests phases) of the proposed algorithm NP-NN are presented in Fig. \ref{fig:large_dataset_conv}.

\begin{table*}[t!]
	\centering
	\includegraphics[width=\linewidth]{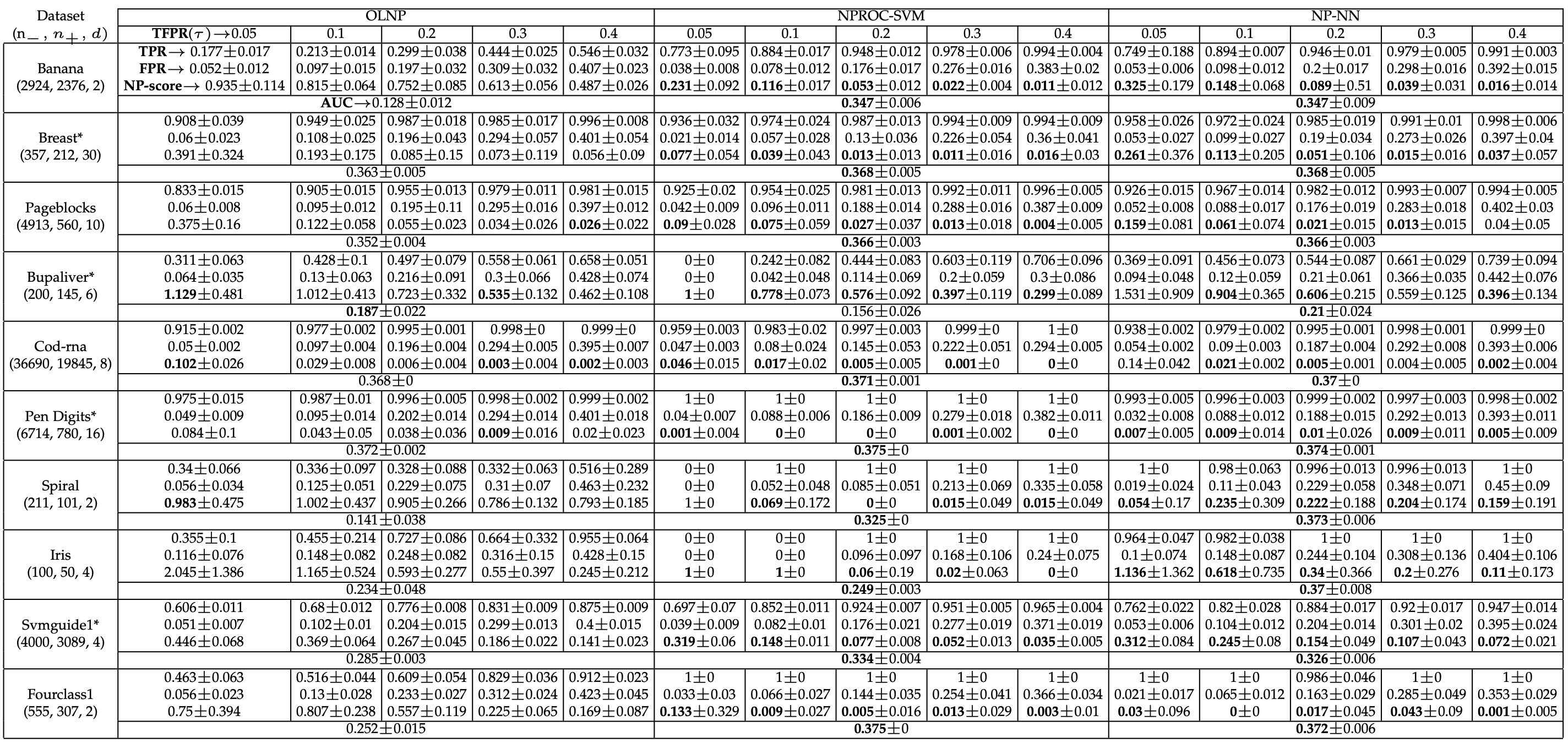}
	\caption{We present performance results of the proposed algorithm NP-NN and the competing algorithms OLNP and NPROC-SVM for each targeted false positive rate (TFPR $\in \{0.05,0.1,0.2,0.3,0.4\}$) on the datasets in the leftmost column. In each case, we run the algorithms $15$ times over $15$ different permutations of the dataset, where $75\%$ ($25\%$) is used for training (testing). Therefore, average of the test results are presented with the corresponding standard deviation. For each dataset, the first row is the achieved true positive rate (TPR), the second row is the achieved false positive rate (FPR), the third row is the NP-score, and the fourth row is the area under curve (AUC) of the receiver operating characteristics (ROC) curve of TPR vs TFPR. The best and the second best performing algorithms are signified with fonts in bold style. Overall, we observe that A) the proposed NP-NN and NPROC-SVM outperform (due to their nonlinear modeling) the other two (including NPROC-LOG based on its results in Fig. \ref{fig:all_rocs}), B) the proposed NP-NN and NPROC-SVM perform comparably in terms of AUC, and NPROC-SVM performs better in terms of NP-score, where the advantage of NPROC-SVM in terms of NP-score seems to disappear as the data size and/or TFPR increase, and C) the proposed NP-NN with online and real-time processing capabilities at $O(N)$ computational and negligible $O(1)$ space complexity has huge advantages over the competing NPROC-SVM in the case of contemporary large scale fast streaming data applications. Namely, when one has a fast streaming dataset of size in the order of millions or more, then the only choice of high performance in real time is the proposed NP-NN, cf. Fig. \ref{fig:large_dataset_conv}.}
	\label{tab:NP_scores_and_AUC}
\end{table*}

The rbf kernel bandwidth parameter $g$ (for the proposed NP-NN as well as NPROC-SVM), the error cost parameter $C$ (for NPROC-SVM) and the number $2D$ of hidden nodes (for the SLFN in the proposed NP-NN) are all $3$-fold cross-validated (based on NP-score) for each random permutation using the corresponding training sequence by a grid search with $g\in\{0.01, 0.02, 0.05, 0.1,$ $0.2, 0.5, 1, 2, 5, 10\}$, $C\in\{0.1, 1, 2, 4\}$ and $D\in\{2, 5, 10, 20, 40, 80, 100\}\times d$, where $d$ is the data dimension. As the regularization has been observed to help little, we opt to use $\lambda\sim 0$ along with SGD learning updates $\eta_t=0.01$ and $0.1 \geq \frac{\beta_t}{\eta_t}\geq 0.01$, randomly initialized $\boldsymbol{w}_1$ and $b_1$ (around $0$) and $\gamma_1=1$ for both the proposed NP-NN and OLNP uniformly in all of our experiments. We directly use the code provided by the authors \cite{tong2018neyman} for NPROC-LOG and NPROC-SVM and also optimize it by the aforementioned cross validation in terms of parameter selection. We observe that for the datasets of relatively short length, algorithms using SGD optimization, i.e., OLNP and NP-NN, improve with multiple passes over the training sequence. Hence, the length of the training sequence of each random permutation is increased by concatenation with additional randomizations for only OLNP and NP-NN (not for NPROC-LOG and NPROC-SVM) during training of both the cross-validation and actual training, resulting in an epoch-by-epoch training procedure. This concatenation is only for training purposes, and hence it is not used in testing and validation, i.e., the actual data size is used in all types of testing to avoid statistical bias and multiple counting.  Our proposed algorithm NP-NN does certainly not need such a concatenation approach for data augmentation in the targeted fast streaming data applications (cf. Fig. \ref{fig:large_dataset_conv}), where data is already abundant and scarcity is not an issue.

\begin{figure*}[t!] 
    \centering
        \includegraphics[width=\linewidth]{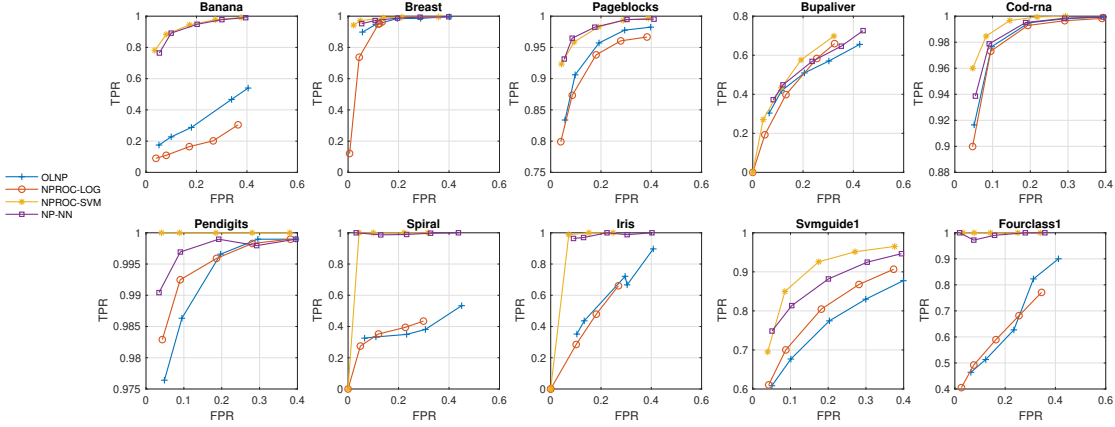}
    \caption{Visual presentations of the results in Table \ref{tab:NP_scores_and_AUC} are provided via the receiver operating characteristics (ROC) curves for all compared algorithms of OLNP, NPROC-LOG, NPROC-SVM and the proposed NP-NN, based on the achieved true positive rates and achieved false positive rates, i.e., TPR vs FPR, corresponding to the targeted false positive rates $\text{TFPR} \in \{0.05, 0.1, 0.2, 0.3, 0.4\}$. Note that the presented ROC curves (TPR vs FPR) are mean curves over $15$ trials of random data permutations for which the standard deviations can be followed from Table \ref{tab:NP_scores_and_AUC}. Overall, in terms of the area under ROC (AUC), we observe that the proposed NP-NN and NPROC-SVM (due to their nonlinear modeling) outperform the other two. On the other hand, the proposed NP-NN performs similarly with NPROC-SVM while providing significant computational advantages. For the quantification of the false positive rate tractability alone, AUC alone and both as a combined measure, we refer to the results in Fig. \ref{fig:banana_dec_boun}, the AUC scores in Table \ref{tab:NP_scores_and_AUC} and the NP-scores in Table \ref{tab:NP_scores_and_AUC}, respectively.}
    \label{fig:all_rocs}
    \vspace{1pt}
\end{figure*}

\begin{figure*}[t!]
    \centering
        \includegraphics[width=\linewidth]{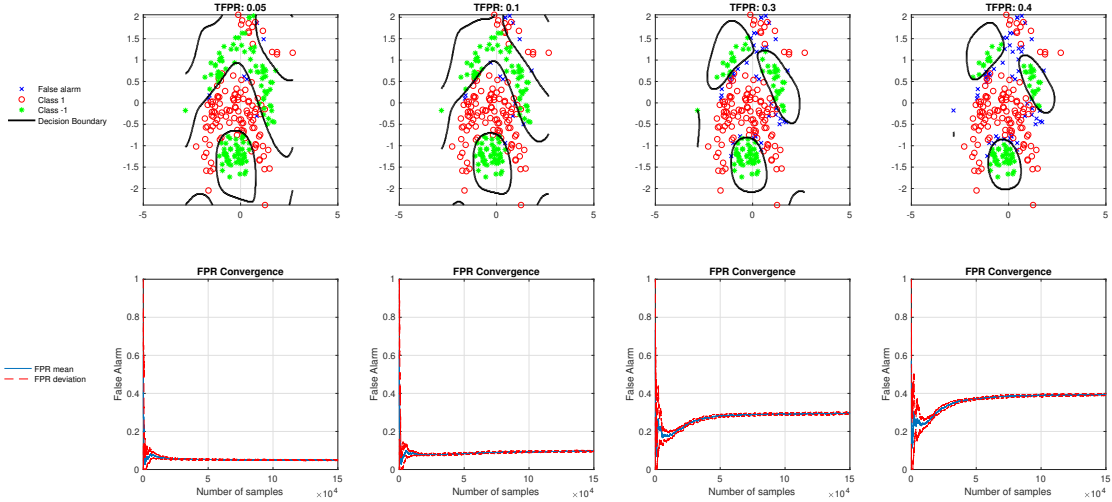}
    \caption{Using the visually presentable $2$-dimensional Banana dataset, upper graphs show the variation in the decision boundary of the proposed NP-NN as the target false positive rate (TFPR) changes as $\text{TFPR} \in \{0.05, 0.1, 0.3, 0.4\}$, and the lower graphs show the mean convergence as well as the standard deviation of the achieved false positive rate (TPR) of the proposed NP-NN over $15$ trials of random data permutations with respect to the number of processed instances during training. To better show the convergence, in each trial, length of the training sequence is increased with concatenation resulting in an epoch-by-epoch training of multiple passes. Overall, as indicated by these results, we observe a decent nonlinear modeling as well as a decent false positive rate controllability with the proposed NP-NN.}
    \label{fig:banana_dec_boun}
\end{figure*}

We run all the algorithms on the test sequence of each of the $15$ random permutations (after training on the corresponding training sequences), and record in each case the achieved false positive rate, i.e., FPR, and true detection rate, i.e., TPR, for the target false positive rates (TFPR) $\tau \in \{0.05, 0.1, 0.2, 0.3, 0.4\}$. For performance evaluation, we compare the mean area under curve (AUC) of the resulting $15$ receiver operating characteristics (ROC) curves of TFPR vs TPR, as well as the mean of the resulting $15$ NP-scores \cite{scott2007performance}, cf. \eqref{eq:NP_score_kappa} with $\kappa=1/\tau$. Note that the mean AUC (higher is better) accounts only for the resulting detection power without regard to false positive rate tractability, whereas the mean NP-score (lower is better) provides an overall combined measure. We evaluate with the both (Table \ref{tab:NP_scores_and_AUC}) in addition to visual presentation of the mean ROC curves (Fig. \ref{fig:all_rocs}) of FPR and TPR. Table \ref{tab:NP_scores_and_AUC} additionally reports the mean TPRs and mean FPRs. We also provide the decision boundaries and the mean convergence of the achieved false positive rate during training for the visually presentable $2$-dimensional Banana dataset (Fig. \ref{fig:banana_dec_boun}). All of our results are provided with the corresponding standard deviations.

We exclude the results of NPROC-LOG in Table \ref{tab:NP_scores_and_AUC} (instead we keep NPROC-SVM since it generally performs better than NPROC-LOG) due to the page limitation, as the table gets too wide otherwise. One can access the results of NPROC-LOG from our Fig. \ref{fig:all_rocs}. Based on our detailed analysis presented in Table \ref{tab:NP_scores_and_AUC} along with the visualization with ROC curves in Fig. \ref{fig:all_rocs}, we first conclude that in general the algorithms NPROC-SVM and the proposed NP-NN with powerful nonlinear classification capabilities significantly outperform the linear algorithms OLNP and NPROC-LOG in terms of both AUC and NP-score, hence the proposed NP-NN and NPROC-SVM better address the need for modeling complex decision boundaries in the contemporary applications. This significant performance difference in favor of nonlinear algorithms NPROC-SVM and the proposed NP-NN is much more clear (especially in terms of the AUC) in highly nonlinear datasets such as Banana, Spiral, Iris, SVMguide1 and Fourclass, as shown in Fig. \ref{fig:all_rocs}. In the case of a small size dataset that seems linear or less nonlinear (e.g., Bupaliver), although OLNP and NPROC-LOG are both linear by design and targeting this dataset with the right complexity and hence expected to be less affected by overfitting, the proposed NP-NN competes with the both well and even slightly outperforms them in terms of AUC (while staying comparable in terms of NP-score). We consider that this is most probably due to the successful compactification of the SLFN in the proposed NP-NN which reduces the parameter complexity by learning the Fourier features in the hidden layer.

As for the comparison between the nonlinear algorithms NPROC-SVM and the proposed NP-NN, we first strongly emphasize that NPROC-SVM has computational complexity (in the worst case of full number of support vectors) between $O(N^2)$ and $O(N^3)$ in training and $O(N)$ in test, where the space complexity is $O(N)$. On the other hand, the proposed NP-NN is truly online without separate training or test phases, which only requires $O(N)$ computational and $O(1)$ negligible space complexity. Hence, NPROC-SVM cannot be applied in our targeted large scale data processing applications due to its prohibitive complexity; nevertheless, we opt to include it in our experiments to set a baseline that is achievable by batch processing. According to the numeric results in Table \ref{tab:NP_scores_and_AUC} and the ROC curves in Fig. \ref{fig:all_rocs}, we first observe that NPROC-SVM and the proposed NP-NN perform comparably in terms of the AUC, hence our technique (thanks to its computationally highly efficient implementation) can be used in large scale applications (where NPROC-SVM computationally fails) without a loss in classification performance. In addition, our algorithm NP-NN outperforms NPROC-SVM in terms of AUC in $3$ datasets; and for small target false positive rate ($\tau=0.05$), the proposed NP-NN has higher TPR compared to NPROC-SVM in $6$ datasets. On the other hand, comparing in terms of the NP-score, NPROC-SVM performs better as a result of enhanced false positive rate controllability due to batch processing. However, this advantage of NPROC-SVM over the proposed NP-NN seems to disappear or decrease as the data size (relative to the dimension) and/or the desired false positive rate increases as observed in the cases of, for instance, Banana and Cod-rna datasets. Therefore, we expect no loss (compared to NPROC-SVM) with the proposed NP-NN in terms of false positive rate controllability as well, when data size increases as in the targeted scenario of the big data applications where NPROC-SVM cannot be used. Indeed, we observe a decent nonlinear classification performance and false positive rate controllability with the proposed NP-NN on, for example,
the Banana dataset ($5300$ instances in only $2$ dimensions), as clearly visualized in Fig. \ref{fig:banana_dec_boun} which shows the false positive rate convergence as well as the nonlinear decision boundaries for various desired false positive rates.    Lastly, NPROC-SVM seems to be failing when TFPR requires only a few mistakes in the non-target class. In this case, NPROC-SVM picks zero mistake resulting in zero TPR and a poor NP-score in return. In contrast, the propsed NP-NN successfully handles such situations as demonstrated by, for instance, TFPR=$0.05$ at Iris dataset in Table \ref{tab:NP_scores_and_AUC}.

Our experiments in Table \ref{tab:NP_scores_and_AUC}, Fig. \ref{fig:all_rocs} and Fig. \ref{fig:banana_dec_boun} include comparisons of the proposed NP-NN with certain batch processing techniques (NPROC-SVM and NPROC-LOG). Hence, we utilize separate training and test phases, along with multiple passes over training sequences (due to small sized datasets in certain cases such as Iris), in those experiments for statistical fairness. However, we emphasize that in the targeted scenario of large scale data applications: 1) one can only use computationally scalable online (such as the proposed NP-NN and OLNP) algorithms, 2)  multiple passes are not necessary as the data is abundant, and also 3) one can target for even smaller false positive rates such as $0.01$ and $0.005$. Therefore, to better address this scenario of large scale data streaming conditions, we conduct additional experiments to compare the online methods (OLNP and the proposed NP-NN) when processing $2$ large datasets (after z-score normalization and $15$ random permutations) on the fly without separate training and test phases based on just a single pass: covertype ($581012$ instances in $54$ dimensions) and Cod-rna ($488565$ instances in $8$ dimensions, this is the original full scale, for which we previously use in Table \ref{tab:NP_scores_and_AUC} a relatively small subset for testing the batch algorithms). We run for $\tau \text{ (TFPR) } \in \{0.005, 0.01\}$ and present the resulting TPR and FPR at each time (in a time-accumulated manner after averaging over $15$ random permutations) in Fig. \ref{fig:large_dataset_conv}. Parameters are set with manual inspection based on a small fraction of the data.

Although the false positive rate constraint is set harder (i.e. smaller as $\tau \text{ (TFPR) } \in \{0.005, 0.01\}$) in this experiment (compared to the smallest TFPR value $0.05$ in Table \ref{tab:NP_scores_and_AUC}), both techniques (OLNP and the proposed NP-NN) successfully converge (the proposed NP-NN appears to converge slightly better) to the target rate (FPR $\rightarrow$ TFPR) uniformly in all cases. Therefore, both techniques promise decent false positive rate controllability (almost perfect) when the data is sufficient. On the other hand, the proposed NP-NN strongly outperforms OLNP in terms of the TPR (again uniformly in all cases), which proves the gain due to nonlinear modeling in the proposed NP-NN. In terms of the NP-score, the proposed NP-NN again strongly outperforms OLNP (except one case, where we observe comparable results). We finally emphasize that the proposed NP-NN achieves this high performance while processing data on the fly in a computation- as well as space-wise extremely efficient manner, in contrast to failing batch techniques in large scale streaming applications due to complexity and failing linear techniques due to insufficient modeling power.

\section{Conclusion}\label{sec:Conclusion}
We considered binary classification with particular regard to i) a user defined constraint on the type I error (false positive) rate that requires false positive rate (FPR) controllability, ii) nonlinear modeling of complex decision boundaries, and iii) computational scalability to voluminous data with online processing. To this end, we propose a computationally highly efficient online algorithm to determine the pair of asymmetrical type I and type II error costs to satisfy the FPR constraint and solve the resulting cost sensitive nonlinear classification problem in the non-convex sequential optimization framework of neural networks. The proposed algorithm is essentially a Neyman-Pearson classifier, which is based on a single hidden layer feed forward neural network (SLFN) with decent nonlinear classification capability thanks to its kernel inspired hidden layer. The SLFN that we use for Neyman-Pearson classification is compact in principle for two reasons. First, the hidden layer exploits -during initialization- the exponential convergence of the inner products of random Fourier features to the true kernel value with sinusoidal activation. Second, learning of the hidden layer parameters, i.e., Fourier features, help to improve the randomly initialized Fourier features. Consequently, the required number of hidden nodes, i.e., the required number of network parameters and Fourier features, can be chosen relatively small. This reduces the parameter complexity and thus mitigates overfitting while significantly reducing the computational as well as space complexity.  Then the output layer follows as a perceptron with identity activation. We sequentially learn the SLFN parameters through stochastic gradient descent based on a Lagrangian non-convex optimization to goal of Neyman-Pearson classification. This procedure minimizes the type II error rate about the user specified type I error rate, while producing classification decisions in the run time. Overall, the proposed algorithm is truly online and appropriate for contemporary fast streaming data applications with real time processing and FPR controllability requirements. Our online algorithm was experimentally observed to either outperform (in terms of the detection power and false positive rate controllability) the state-of-the-art competing techniques with a comparable processing and space complexity, or perform comparably with the batch processing techniques, i.e., not online, that are -however- computationally prohibitively complex and not scalable.

\bibliography{mybibfile}

\end{document}